# ANALYZING ARCHITECTURES FOR NEURAL MACHINE TRANSLATION USING LOW COMPUTATIONAL RESOURCES

Aditya Mandke[*], Onkar Litake[*], and Dipali Kadam[*]

Department of Computer Engineering, SCTR's Pune Institute of Computer Technology, Pune, Maharashtra, India

## ABSTRACT

*With the recent developments in the field of Natural Language Processing, there has been a rise in the use of different architectures for Neural Machine Translation. Transformer architectures are used to achieve state-of-the-art accuracy, but they are very computationally expensive to train. Everyone cannot have such setups consisting of high-end GPUs and other resources. We train our models on low computational resources and investigate the results. As expected, transformers outperformed other architectures, but there were some surprising results. Transformers consisting of more encoders and decoders took more time to train but had fewer BLEU scores. LSTM performed well in the experiment and took comparatively less time to train than transformers, making it suitable to use in situations having time constraints.*

## 1. INTRODUCTION

Machine Translation (MT) can be used to automatically convert a sentence in one (source) language into its equivalent translation in another language (target) using Natural Language Processing. Earlier, rule-based MT was the norm, with heuristic learning of phrase translations from word-based alignments and lexical weighting of phrase translations [1].

Later, seq2seq networks [2,3,4] have dominated the field of Machine Translation by replacing statistical phrase-based methods. This led to a corpus-based MT system. The encoder-decoder paradigm has risen with Neural Machine Translation (NMT) [5]. For this purpose, both the encoder and decoder were implemented as Recurrent Neural Networks. This helps to encode a variable-length source sentence into a fixed-length vector and the same is decoded to generate the target sentence. For this, Long Short Term Memory (LSTM) units [6,7] were utilized to improve the translation quality of the target sentence. The RNN based NMT approach quickly became the norm and was utilized in industrial applications [8, 9, 10].

Later, convolutional encoders and decoders [11] were introduced, which allows encoding the source sentence simultaneously compared to recurrent networks for which computation is constrained by temporal dependencies. Conv2Seq architectures [12] utilized this to fully parallelize the training and optimization process over GPUs and TPUs, leading to faster training speed.

Most recently, by utilizing self-attention and feed-forward connections, Transformers [13] further advanced the field of Translation by increasing the efficiency and reducing the speed of convergence. One of the downsides of using a Transformers-based architecture is that it requires larger training data for convergence than an RNN-based model.

[*]Equal Contribution

seq2seq networks can also be used to train multi-way, multilingual neural machine translation (MNMT) models [14, 15]. This approach enables a single translation model to translate between multiple languages, which is made possible by having a single attention mechanism that is shared across all language pairs.

[16] describes that advances in machine translation have led to a steep increase in performance on tasks for many high resource languages [17] Since translation is a resource-hungry task, it works well in the case of high-resource languages [18]. In comparison to their European counterparts, languages of the Indian subcontinent (Indic Languages) don't have large-scale sentence-aligned corpora.

In comparison with high-resource languages, there have been very few efforts to benchmark NMT models on low-resource Indic Languages. In our approach, we try to evaluate a variety of popular architectures used in the field of Translation on the Indic language "Marathi". Marathi ranks 15th in the world, with about 95 million speakers. For the dataset, we use a subset of Samantar corpora which has nearly 3.3 million English-Marathi parallel sentences. We observe that transformers outperform both RNN-based, and Conv2Seq-based architectures in terms of accuracy, but take a much larger time to train per epoch.

For most of the research work, large models requiring heavy computational resources are trained for a long duration of time. For example, 38 hours of training on 4 A100 GPUs were used for training of IndicTrans on the Samantar dataset. Most of the time it's not possible to have such a computationally high setup. We are training models with low computational resources and investigating their work. We are using the NVIDIA TITAN X GPU for training our machine translation models. We also do a comparison between the BLEU scores and the time required to train the model to find which architecture is suitable for different scenarios.

The rest of the paper is structured as follows. Section 2 surveys the related work in the field of Neural Machine Translation. Section 3 indicates the System description which includes the dataset we used, the preprocessing tasks, and our training procedure. Section 4 displays the different architectures we used for training the model. Section 5 displays our results and the subsequent section 6 states the conclusions drawn from the results. Section 7 states the future work.

## 2. RELATED WORK

[5] in 2014 conjectured that the usage of a fixed-length vector is a performance limitation in this basic encoder-decoder architecture. They proposed a model for automatically (soft-)searching for portions of a source sentence that are useful for predicting a target word without having to explicitly create these parts as a hard segment.

[9] in 2016 introduced a deep LSTM network with 8 encoder and decoder layers having attention and residual connections. In the attention mechanism, the bottom layer of the decoder was connected to the top layer of the encoder to improve parallelism and reduce training time. Wordpieces for both input and output were introduced. It helped to reduce the average error in translation by 60%.

[19] in 2018 identified several key modeling and training techniques and applied them to RNN architectures which resulted in the emergence of a new RMNT + model that outperformed three fundamental architectures on the benchmark WMT '14 English to French and English to German tasks. A new hybrid architecture was introduced for Neural Machine Translation.

[20] looks into the Marathi-English translation task by taking into consideration five architectures, namely Google, wmt-en-de, iwslt-de-en, wmt-en-de-big-t2t, vaswani-wmt-en-de-big. The corpora considered here was the Tatoeba, Wikimedia, and bible dataset, and also data gathered through web-scraping, leading to about 3 million sentences.

## 3. System Description

### 3.1. Dataset

For supervised Neural Machine Translation, we generally need a large number of parallel corpora between the two languages from a variety of domains. The Samantar dataset [21] has nearly 3.3 million English-Marathi parallel sentences mined from a variety of sources such as JW300 [22], cvit-pib [23], wikimatrix [24], kde4, pmindia v1 [25], gnome, bible-uedin [26], ubuntu, ted2020 [27], mozilla-I10nm Tatoeba, tico19 [28], ELRC_2922, IndicParCorp, Wikipedia, PIB_Archives, Nouns_Dictionary, NPTEL, Timesofindia, and Khan_academy. It contains a total of 3288874 sentences, of which the first 1000000 were chosen.

Table 1. Samantar dataset aggregation description

| Existing dataset | Corpora size (in thousands) |
|---|---|
| JW300 | 289 |
| cvit-pib | 114 |
| wikimatrix | 124 |
| KDE4 | 12 |
| PMIndia V1 | 29 |
| GNOME | 26 |
| bible-uedin | 60 |
| Ubuntu | 26 |
| TED2020 | 22 |
| Mozilla-110n | 15 |
| Tatoeba | 53 |
| tico19 | < 1 |
| IndicParCorp | 2600 |
| Wikipedia | 24 |
| PIB | 74 |
| PIB_archives | 29 |
| Nouns_dictionary | 57 |
| NPTEL | 15 |
| Timesofindia | 25 |
| Khan_academy | < 1 |

## 3.2. Preprocessing

For initial preprocessing, we used the scripts available along with indicTrans [21]. To handle the problems of OOV (out-of-vocabulary) and rare words, subword segmentation techniques were applied [29]. If a word encountered is an unknown word, still, the model can translate the word accurately by treating it as a sequence of subword units. For doing this, multiple techniques such as Byte Pair Encoding, Unigram Language Model, Subword Sampling, BPE-dropout can be used. In our experiments, we use Byte Pair Encoding.

Byte Pair Encoding is a word segmentation technique that uses a compression algorithm. Splitting words into sequences of characters iteratively combines the most frequent character pair into one. BPE can represent an open vocabulary through a fixed-size vocabulary of variable-length character sequences. Thus, it is a very useful method for word segmentation in Neural Network Models.

## 3.3 Training and Evaluation

We use fairseq [30] for this purpose. Before feeding data into the model, we binarized it for faster loading. The model was trained on the NVIDIA TITAN X GPU. To mimic training on a larger number of GPUs, we used the update-freq parameter. As the optimizer, Adam [31] was used, with the Adam betas sent to 0.9 and 0.98. The loss function used was cross-entropy. For regularization, label smoothing [32] of 0.1 and a dropout [33] of 0.3 were applied. A weight decay of $10^{-4}$ was used and the learning rate scheduler used was inverse_sqrt. While evaluating the accuracy of a translation with BLEU [34], the sacrebleu library was used.

For evaluation, we used the BLEU metric as made available in the SacreBleu library, and to make it fair it was done on the 25th iteration of the models. Since it was being done on processed text, the byte pair encoding components had to be removed. A Beam Search of 5 was utilized for generating the text.

For evaluating translation, the aspects of fluency, adequacy, Human-mediated Translation Edit rate, and Fidelity need to be considered. Human evaluation, the benchmark method, is time-consuming and expensive. Thus, the quick, inexpensive, and language-independent BLEU [34] metric was proposed. It is a score that ranges from 0 to 1, where a score of 1 means that the candidate machine translation is identical to one of the reference translations [35].

## 4. ARCHITECTURES USED

We trained our dataset on a total of 5 different architectures as follows.

### 4.1. fconv_iwslt_de_en

It is entirely based on convolutional networks. Compared to recurrent models, computations over all elements can be fully parallelized during training to better exploit the GPU hardware and optimization is easier since the number of non-linearities is fixed and independent of the input length [11, 12].

### 4.2. lstm

It is a vanilla RNN architecture that uses LSTMs as its encoder and decoder. LSTMs were designed to overcome the long-term dependency problem faced by RNNs and mitigate short-term memory using mechanisms called gates. In our experiments, the encoder embedding dimension of the LSTM is 256, and the decoder embedding dimension is 512 [6, 7].

### 4.3 lstm_wiseman_iwslt_de_en

It is an LSTM but has smaller embedding dimensions. Also known as a tiny-lstm, it has an encoder embedding dimension of 256, and a decoder embedding dimension of 256.

### 4.4 transformer_iwslt_de_en

The Transformer is the first model which converts input sequences to output sequences relying entirely on self-attention to compute representations of its input and output without using sequence-aligned RNNs or convolution. Having 6 encoder and decoder layers with an embedding dimension of 512, it utilizes self-attention with 8 encoder and decoder attention heads [13].

### 4.5 transformer_wmt_en_de_big_t2t

This is also a transformer model like the one described in (4.4) but has a bigger size, having 16 encoder and decoder attention heads, and the encoders and decoders have an embedding dimension of 1024. It also utilizes attention dropout and activation dropout.

## 5. RESULTS

Table 2. Best BLEU scores & Training time in hours

| Architecture | Best BLEU | Time (in hours) |
|---|---|---|
| transformer_iwslt_de_en | 22.98 | 11 |
| transformer_wmt_en_de_big_t2t | 21.17 | 16.5 |
| lstm | 17.24 | 6.25 |
| fconv_iwslt_de_en | 15.41 | 5.5 |
| lstm_wiseman_iwslt_de_en | 14.03 | 4 |

Surprisingly transformer_iwslt_de_en outperformed transformer_wmt_en_de_big_t2t. Transformer_iwslt_de_en had a lesser number of encoders, decoders and attention heads compared to transformer_wmt_en_de_big_t2t. Having more encoders and decoders, transformer_wmt_en_de_big_t2t took significantly more time to train. This shows that transformer_wmt_en_de_big_t2t is a better model to train in the given circumstances for machine translation as it has more BLEU scores and takes less time to train. One noticeable thing obtained from the results is the performance of LSTM. Though its BLEU score is 17.24 and is less than transformer_wmt_en_de_big_t2t but the training time required is significantly very low. If having time constraints, the LSTM model can be used for Machine translation tasks as it gives a high BLEU score in less time.

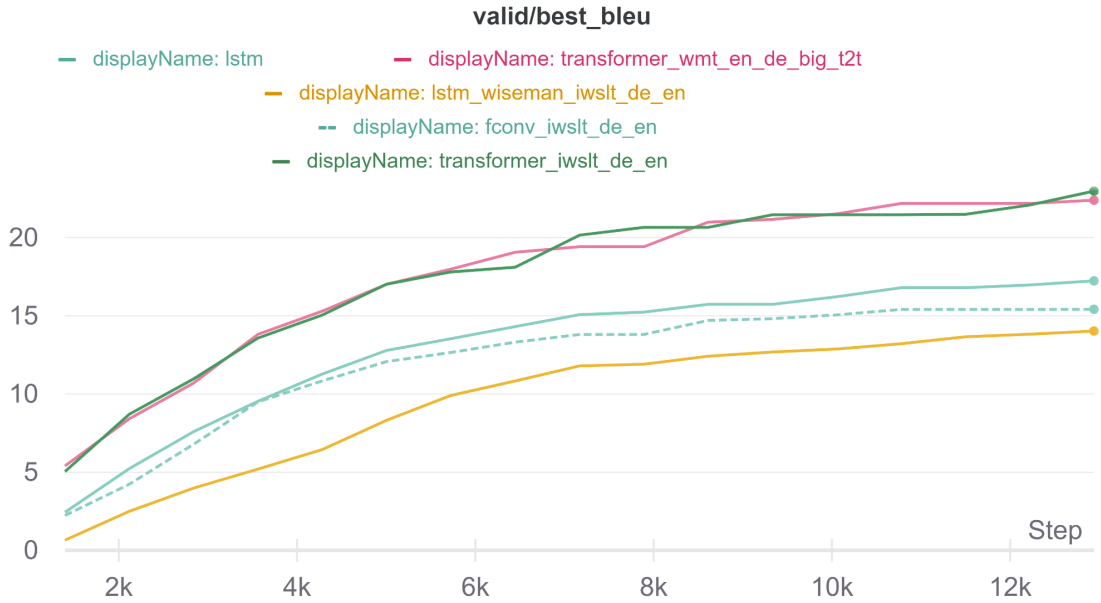

Figure 1. The Best BLEU scores on the Valid dataset against time steps

Figure 1 shows the Best BLEU scores per architecture over time. It can be observed that the trend of optimization is similar, with the rate of change of the score beginning to reduce at around 8000 steps. We used Weights & Biases [36] for experiment tracking and visualizations to develop insights for this paper.

Following are a few examples of sentences and their translated text which were inferred from the model transformer_iwslt_de_en:

Table 3. Translation Results

| Marathi(source language) | English(destination language) |
|---|---|
| तो एक व्यापारी आहे. | He is a businessman. |
| आज खूप गरम आहे. | It is very hot today. |
| नदी ओसंडून वाहत आहे. | The river is overflowing |
| शरद पवार हे देशातील सर्वात प्रभावी राजकारण्यांपैकी एक आहेत. | Sharad Pawar is one of the most influential leaders of the country. |

## 6. CONCLUSION

The study of different architectures for machine translation helps us to understand which model can perform best with low computational resources. In this work, we looked into a multitude of neural network architectures, such as LSTMs, Conv2Seq networks, and Transformers. For our task, Transformers gave the best performance with respect to the metric BLEU as compared to the remaining architectures. But this comes at a large computational cost. Transformers require nearly 2-3x time per epoch as compared with other networks.

It was observed that even though transformer_wmt_en_de_big_t2t is larger in size as compared with transformer_iwslt_de_en, it was not able to give better accuracy. LSTMs were able to show a good performance in less time as compared with Transformers. Depending upon the scenario such as availability of computational resources, time required to train a model, different architectures can be chosen. Even with low computational resources and a comparatively smaller corpora, good results can be achieved using the Transformer architecture.

## 7. FUTURE WORK

For this work, we have trained the models on a subset of the dataset. This was done as a result of a limitation of computational limitations. In the future, we plan to train the model on the whole dataset instead of the subset of the dataset. We plan to train these models on a setup having high computational power to investigate the change in BLEU score and required time as compared with low computational resources. We plan to train the model parallelly to detect if it causes any changes to the BLEU score.

In addition to this, we will look into replicating these experiments with different languages (for example English, Hindi, Mandarin etc), while replicating the low-resource setting and see whether the observed results remain the same. In this experiment, for preprocessing, we have used Byte Pair Encoding for treating a word as a sequence of subword units. We plan to use other algorithms such as Unigram Language Model, Subword Sampling, BPE-dropout and see whether the outcomes observed in this experiment persist.

## ACKNOWLEDGEMENTS

We would like to thank the SCTR's PICT Data Science Lab for letting us use the resources for an extended time. We would also like to thank Prof. Mukta Takalikar and Dr. Prahlad Kulkarni for their continuous help and motivation.

Also, we thank the anonymous reviewers for their insightful comments.
## ACKNOWLEDGEMENTS

We would like to thank the SCTR's PICT Data Science Lab for letting us use the resources for an extended time. We would also like to thank Prof. Mukta Takalikar and Dr. Prahlad Kulkarni for their continuous help and motivation.

Also, we thank the anonymous reviewers for their insightful comments.